\documentclass[conference]{IEEEtran}
\IEEEoverridecommandlockouts
\usepackage{cite}
\usepackage{amsmath,amssymb,amsfonts}
\usepackage{algorithmic}
\usepackage{graphicx}
\usepackage{textcomp}
\usepackage{xcolor}
\usepackage{svg}
\usepackage{url}
\usepackage[numbers]{natbib}
\def\BibTeX{{\rm B\kern-.05em{\sc i\kern-.025em b}\kern-.08em
    T\kern-.1667em\lower.7ex\hbox{E}\kern-.125emX}}
\begin{document}

\title{Small Object Detection with YOLO: A Performance Analysis Across Model Versions and Hardware\\
}

\author{\IEEEauthorblockN{Muhammad Fasih Tariq}
\IEEEauthorblockA{\textit{School of Systems and Technology} \\
\textit{University of Management and Technology}\\
fasihtariq2611@gmail.com}
\and
\IEEEauthorblockN{ Muhammad Azeem Javed}
\IEEEauthorblockA{\textit{School of Systems and Technology} \\
\textit{University of Management and Technology}\\
azeem.javeed@umt.edu.pk}
}

\maketitle

\begin{abstract}
This paper provides an extensive evaluation of YOLO object detection models (v5, v8, v9, v10, v11) by comparing their performance across various hardware platforms and optimization libraries. Our study investigates inference speed and detection accuracy on Intel and AMD CPUs using popular libraries such as ONNX and OpenVINO, as well as on GPUs through TensorRT and other GPU-optimized frameworks. Furthermore, we analyze the sensitivity of these YOLO models to object size within the image, examining performance when detecting objects that occupy 1\%, 2.5\%, and 5\% of the total area of the image. By identifying the trade-offs in efficiency, accuracy, and object size adaptability, this paper offers insights for optimal model selection based on specific hardware constraints and detection requirements, aiding practitioners in deploying YOLO models effectively for real-world applications.
\end{abstract}


\section{Introduction}
In recent years, Deep Neural Networks (DNNs) have gained dominance for their performance on datasets that surpass even human accuracy on object classification and localization. As such, people would consider it a "solved" problem; however, this leaves out the key issues with the practicality of these models. That being object size and inference speed. Unlike humans, they face issues when detecting objects on multiple scales and they have a heavy computational cost making real-time detection difficult. Multi-stage detectors such as RCNN \cite{rcnn}, Faster RCNN \cite{fasterrcnn}, Mask RCNN \cite{maskrcnn} can provide improved performance on this task but at a high computational cost, leading to lower frames processed per second and higher. Single Stage Detectors (SSD) such as YOLO \cite{yolov1} and Detectron \cite{Detectron2018} offer speed and throughput for relatively lower accuracy.
This research aims to address this speed, performance issues, and object size performance through a rigorous evaluation of YOLO object detection models, specifically focusing on:
\begin{itemize}
    \item Performance comparison across multiple YOLO versions (v5, v8, v9, v10, v11)
    \item Inference speed assessment using different optimization libraries
    \item Model sensitivity to object size within image contexts
\end{itemize}
Particularly novel in our approach is the detailed investigation of model performance relative to object size, examining the detection accuracy for objects occupying minimal image areas (1\%, 2.5\%, and 5\%). This analysis offers insights into model robustness across different scales of object detection, a critical consideration for practical applications ranging from medical imaging to surveillance systems.

Multiple surveys conducted on small object detection have demonstrated a drop in accuracy metrics in comparison to salient objects within the image for DNNs \cite{surveysmallobjects}\cite{dnnsmallobject}. In \cite{smallobjectsurveywithpillers} they describe common techniques for improving small object DNNs such as multi-scale detection, super-resolution \cite{superresolution} and region proposals. YOLOv3 introduced multiscale prediction heads \cite{yolov3} and further success has been observed by modifying the yolov3 darknet architecture for small objects while maintaining the same FLOPs \cite{yolov3UAV} by enriching the early layers with spatial information. In \cite{yolov5small} they attempted to improve yolov5's performance on small objects by incorporating local FCN and yolov5. In \cite{yolov8attention} Yolov8's performance on small objects was improved via incorporating a sparse attention mechanism. \\
When it comes to the software side of things, different deep learning frameworks offer different benefits and speedups depending on the hardware \cite{inferenceframeworks}. In \cite{yoloembedded} three key optimization strategies are outlined for DNNs: 1) Model architecture, 2) Compression, and 3) Platforms. In this paper, we focus on the third part. NVIDIA's tensorrt provides platform-specific speedups such as kernel fusion \cite{nvidiayolocomp}. \cite{openvino} demonstrated a 141\% improvement in model inference utilizing Intel's openvino platform. Meanwhile, the ONNX platform provides a common platform for the deployment and transfer of models to other frameworks \cite{yoloembedded}.
Our research focuses on providing insight into the performance and speed of these models, using a variety of frameworks for the task of small object detection, and offering a look at how these models can perform utilizing different types of hardware.
\section{Experiments}
\subsection{Experimental Setup}
For our experiments we used Python 3.11.10, Pytorch 2.4.2, Ultralytics 8.3.39, CUDA 12.2. The experiments were performed on an Intel i7-13700 with 16 GB of Ram. For the AMD system, a Ryzen 7 5800X was utilized. For the GPU, we used an NVIDIA RTX 3070. Each experimental run was repeated five times for consistency. All runs were performed on Ubuntu 24.04.1 LTS. All networks were tested in the MS COCO 2017 dataset \cite{mscoco} and to verify performance for small-scale objects, we use the Large-Scale Benchmark and Challenges for Object Detection in Aerial Images (DOTAv1.5) \cite{DOTA}. For the models, we chose the smallest available variants and for yolov5 we used the n6u version. We tested these models at different resolutions and using the following backends: Onnx, PyTorch, Intel Extension for Pytorch (IPEX), Openvino and Tensorrt. The precision tests are carried out on images of 640x640 in size for the COCO dataset and 1024x1024 for the DOTAv1.5 dataset. For training we trained the models for 100 epochs using the AdamW optimizer \cite{AdamW} .
\begin{figure}[htbp]
\centerline{\includegraphics[width=0.5\textwidth]{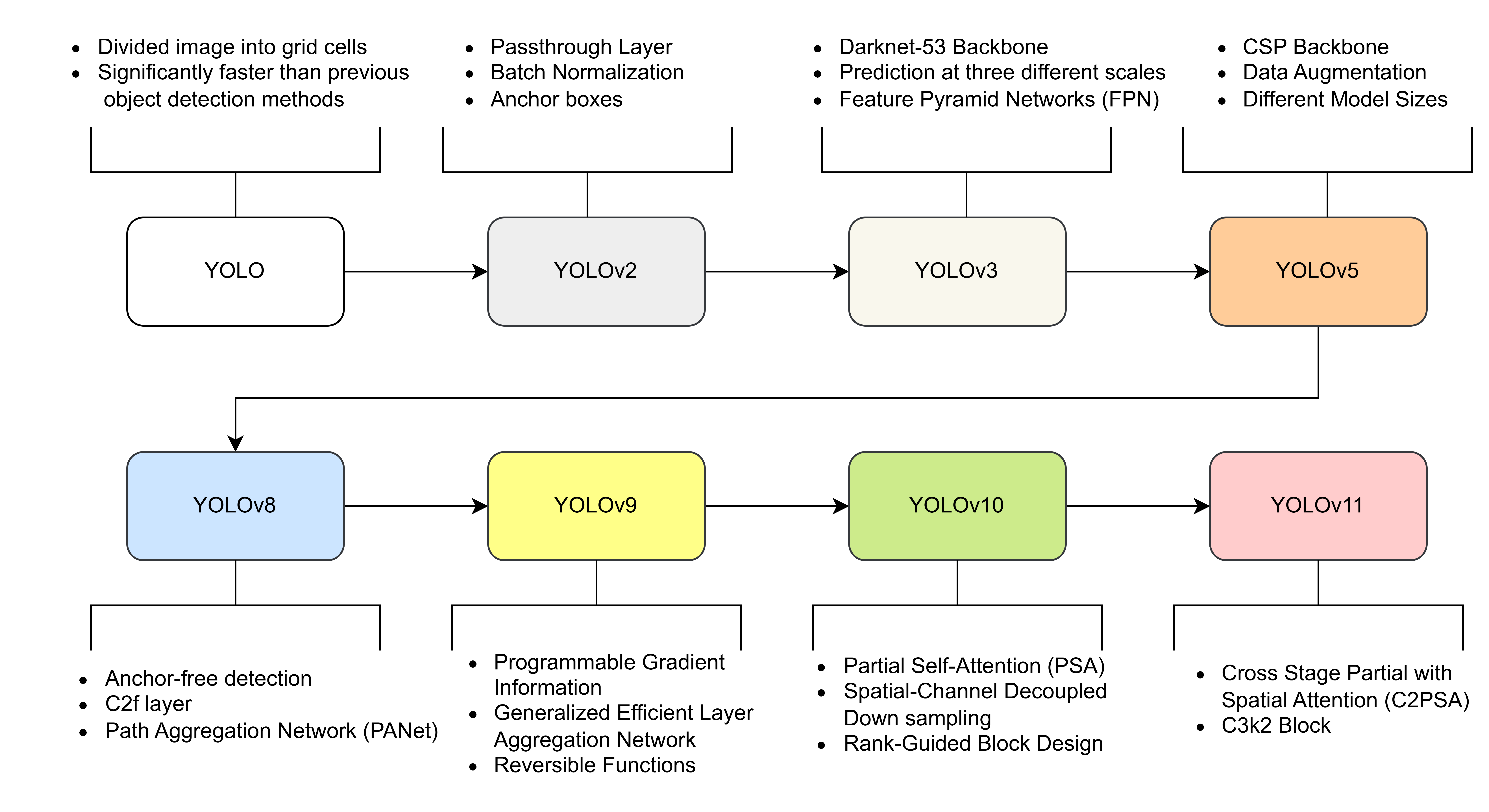}}
\caption{History of the YOLO series.}
\label{yolohistory}
\end{figure}
\subsection{Metrics}
For measuring the accuracy of the models, we used the mean average precision (mAP) metric and its variants such as mAP50 and mAP 0.5:95 as it is common convention.

\begin{equation}
\text{AP}_c = \int_0^1 p_c(r) dr \approx \sum_{i=1}^n (r_{i+1} - r_i)p_c(r_i) 
\end{equation}

\text{where:}
\begin{itemize}
\item $\text{AP}_c$ : Average Precision for class $c$
\item $p_c(r)$ : precision value at recall level $r$ for class $c$
\item $n$ : number of samples
\item $r_i$ : recall value at step $i$
\end{itemize}

\begin{equation}
\text{mAP} = \frac{1}{|C|} \sum_{c=1}^{|C|} \text{AP}_c
\end{equation}

\text{where:}
\begin{itemize}
\item $|C|$ : total number of classes
\item $\text{AP}_c$ : Average Precision for class $c$
\end{itemize}

\begin{equation}    
\text{mAP}_{[.5:.95]} = \frac{1}{10} \sum_{t=0.5}^{0.95} \text{mAP}_t
\end{equation}

\text{where:}
\begin{itemize}
\item $t$ : IoU threshold varying from 0.5 to 0.95 in steps of 0.05
\item $\text{mAP}_t$ : mAP at specific IoU threshold $t$
\end{itemize}

\begin{equation}
\text{Throughput} = \frac{\sum_{i=1}^N (t_{\text{pre},i} + t_{\text{infer},i} + t_{\text{post},i})}{N}
\\
\end{equation}
\begin{itemize}
\item $N$ : total number of images processed
\item $t_{\text{pre},i}$ : pre-processing time for image $i$
\item $t_{\text{infer},i}$ : inference time for image $i$
\item $t_{\text{post},i}$ : post-processing time for image $i$
\end{itemize}

\subsection{Experiment Results}
When comparing the precision of these models, YOLOv11 shows the best performance in the dataset and is comparable to YOLOv10 in mAP, however, for small objects v10 does outperform it by 1.5 mAP. For medium sized objects, v5, v10, v11 show a similar level of precision. At 5\%, YOLOv5 outperforms all other models by a wide margin, as shown by Table \ref{precision table}. It should be pointed out that our mAP(All) values do not mirror the ultralytics published numbers. This is because in their code, mask annotation information is used, which leads to higher performance, but in our view that is not an accurate reflection of their real-world performance, and as such we have chosen to omit mask annotation information when evaluating these models.
\begin{table*}
    \centering
    \caption{All the different precision results at 1\%, 2.5\%, 5\% of the total image size (640x640) for the different YOLO models and their overall performance}
    \label{precision table}
    \begin{tabular}{|c|c|c|c|c|c|c|c|c|} \hline 
        Model & 
        \begin{tabular}[c]{@{}c@{}}mAP@0.5\\(All)\end{tabular} & 
        \begin{tabular}[c]{@{}c@{}}mAP@0.5:0.95\\(All)\end{tabular} & 
        \begin{tabular}[c]{@{}c@{}}mAP@0.5\\(Small)\end{tabular} & 
        \begin{tabular}[c]{@{}c@{}}mAP@0.95\\(Small)\end{tabular} & 
        \begin{tabular}[c]{@{}c@{}}mAP@0.5\\(Medium)\end{tabular} & 
        \begin{tabular}[c]{@{}c@{}}mAP@0.5:0.95\\(Medium)\end{tabular} & 
        \begin{tabular}[c]{@{}c@{}}mAP@0.5\\(Large)\end{tabular} & 
        \begin{tabular}[c]{@{}c@{}}mAP@0.5:0.95\\(Large)\end{tabular} \\ \hline 
        V5n & 48.73 & 35.00 & 42.95 & 30.40 & \textbf{63.04} & \textbf{48.96} & \textbf{74.51} &  \textbf{64.54} \\ \hline 
        V8n & 47.93 & 33.84 & 43.36 & 31.95 & 59.03 & 44.91 & 66.07 & 49.25 \\ \hline 
        V9t & 48.60 & 35.04 & 44.08 & 32.53 & 59.47 & 46.28 & 67.94 & 57.69 \\ \hline 
        V10n & 50.74 & 36.31 & \textbf{48.26} & \textbf{35.37} & 62.46 & 48.28 & 70.46 & 57.62 \\ \hline 
        V11n & \textbf{51.45} & \textbf{36.60} & 47.26 & 33.88 & 63.00 & 48.50 & 71.98 & 57.90 \\ \hline
    \end{tabular}
\end{table*}
\begin{table*}
    \centering
    \caption{All the different precision results for 1024x1024 input size across different YOLO models and their overall performance on the DOTAv1.5 dataset}
    \label{precision table}
    \begin{tabular}{|c|c|c|c|c|c|c|c|c|} \hline 
        Model & 
        \begin{tabular}[c]{@{}c@{}}mAP@0.5\\(All)\end{tabular} & 
        \begin{tabular}[c]{@{}c@{}}mAP@0.5:0.95\\(All)\end{tabular} & 
        \begin{tabular}[c]{@{}c@{}}mAP@0.5\\(Small)\end{tabular} & 
        \begin{tabular}[c]{@{}c@{}}mAP@0.5:0.95\\(Small)\end{tabular} & 
        \begin{tabular}[c]{@{}c@{}}mAP@0.5\\(Medium)\end{tabular} & 
        \begin{tabular}[c]{@{}c@{}}mAP@0.5:0.95\\(Medium)\end{tabular} & 
        \begin{tabular}[c]{@{}c@{}}mAP@0.5\\(Large)\end{tabular} & 
        \begin{tabular}[c]{@{}c@{}}mAP@0.5:0.95\\(Large)\end{tabular} \\ \hline 
        V5n & 43.60 & 26.02 & 61.56 & 40.93 & 55.05 & 40.73 & 51.88 & 27.80 \\ \hline 
        V8n & \textbf{48.33} & \textbf{29.60} & \textbf{67.88} & \textbf{46.46} & 62.58 & 47.74 & \textbf{60.93} & \textbf{35.73} \\ \hline 
        V9t & 46.98 & 28.90 & 61.71 & 42.06 & \textbf{74.01} & \textbf{54.68} & 35.67 & 24.49 \\ \hline 
        V10n & 40.95 & 25.14 & 51.16 & 36.23 & 54.18 & 40.93 & 37.05 & 25.71 \\ \hline 
        V11n & 42.70 & 25.63 & 64.33 & 43.10 & 51.68 & 36.43 & 22.45 & 14.42 \\ \hline
    \end{tabular}
\end{table*}

\begin{figure}
    \centering
    \includegraphics[width=1\linewidth]{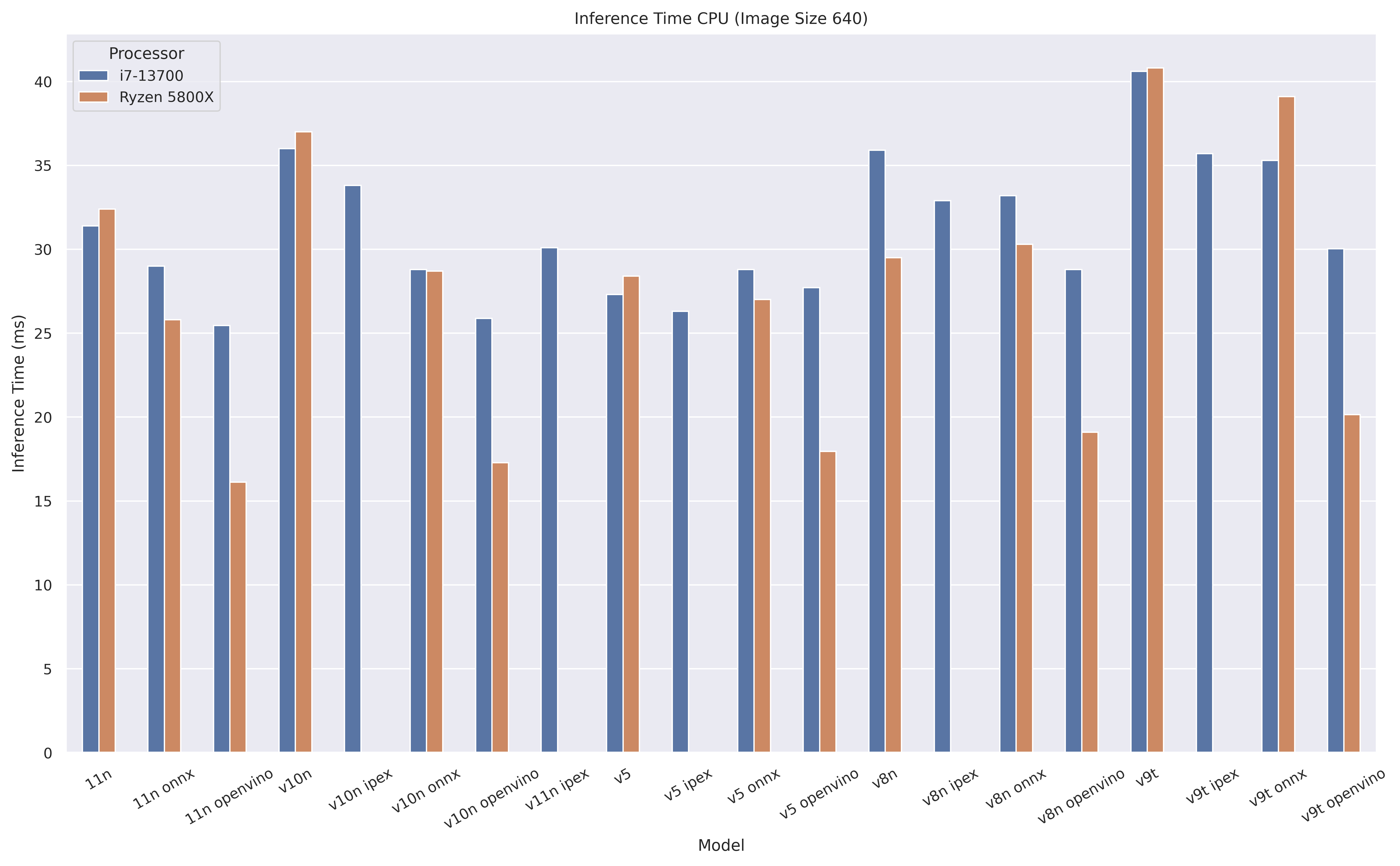}
    \caption{CPU Performance on the AMD \& Intel Systems on multiple inference engines.}
    \label{fig:cpu inference}
\end{figure}

When it comes to performance between these models and their respective software backends. The AMD system despite having a weaker CPU on paper, does outperform the intel system especially when using the openvino framework which shows the best overall performance as shown by figure \ref{fig:cpu inference}. 

\subsection{GPU Performance}
\begin{figure}
    \centering
    \includegraphics[width=1\linewidth]{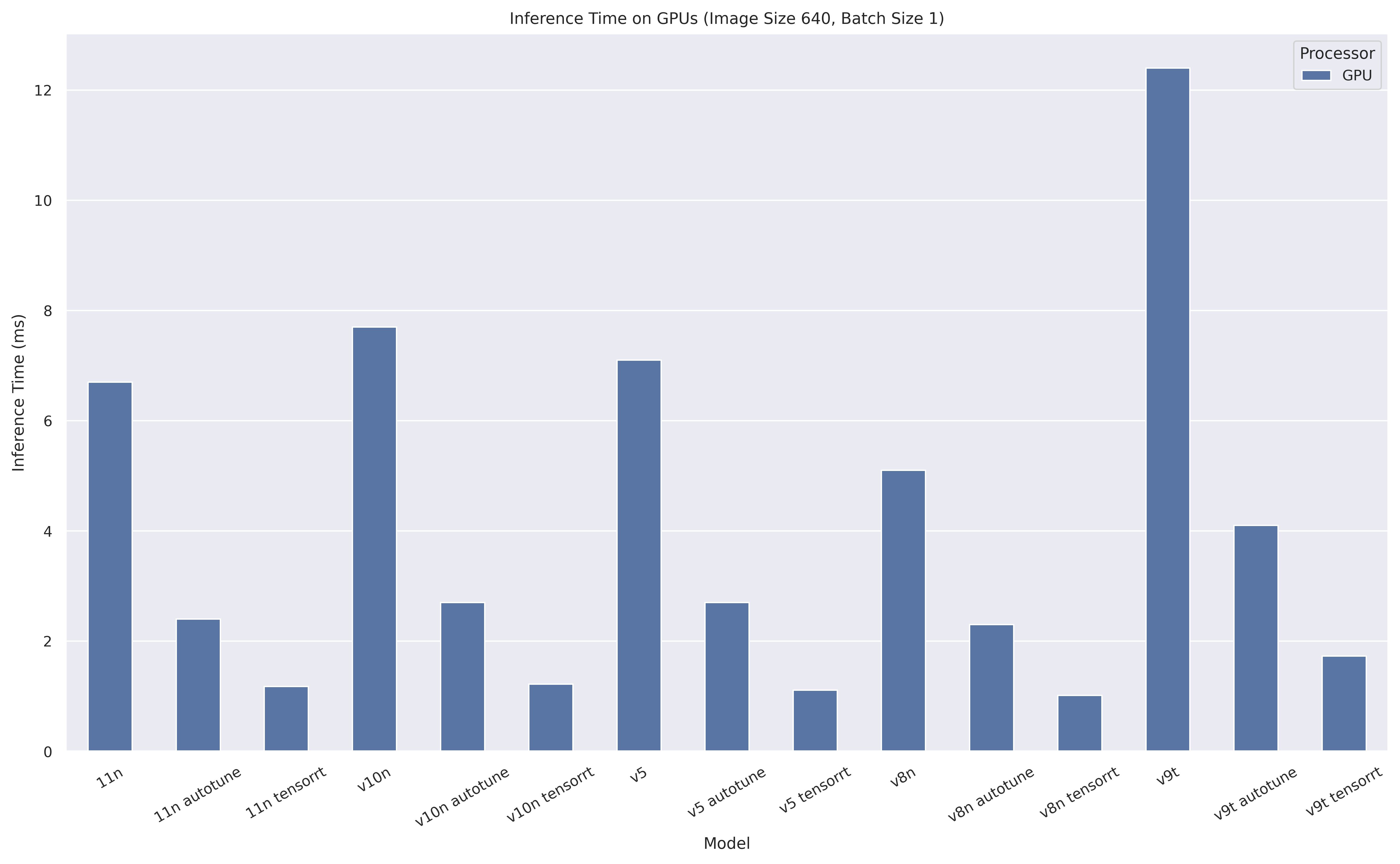}
    \caption{Inference time on RTX 3070 for all models}
    \label{fig:gpu inference}
\end{figure}

Moving onto the GPU performance, Tensorrt is significantly faster than any other format, surpassing them in inference speed by the widest margin so far, this is due to the plans that tensorrt formulates including operations such as kernel fusion. It can also be noted that despite having a higher FLOP count, YOLOv5 has a lower inference time than both v10 and v11 on the tensorrt framework, as shown in figure \ref{fig:gpu inference}
\begin{figure}
    \centering
    \includegraphics[width=1\linewidth]{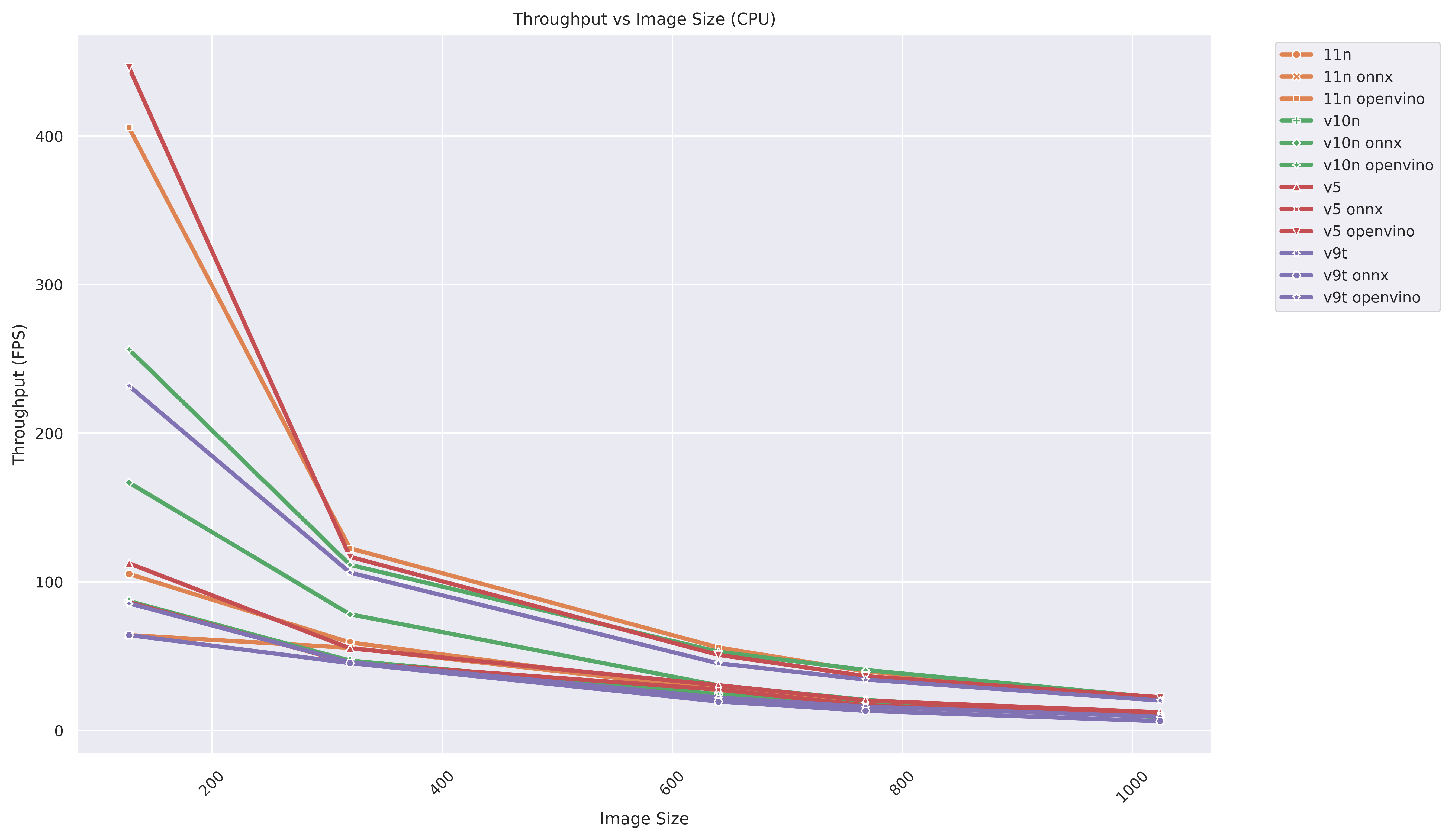}
    \caption{Effect of image size of throughput on CPU}
    \label{fig:imagesize cpu}
\end{figure}
\subsection{Effect of Image Size}
The effect of the image size as expected exponentially reduces the throughput of the models, as the RTX 3070 becomes overwhelmed at higher image sizes and leads to a much sharper drop in throughput beyond the 640x640 image size as shown by Figure \ref{fig:gpu inference}.
However, for the CPU, larger image sizes lead to a consistent sharp drop in throughput beyond the 320x320 image size, as shown by Figure \ref{fig:imagesize cpu}. When it comes to accuracy measurements, an interesting trend emerges. For MS-COCO, both v11 and v5 show a similar level of performance, followed by v10, v9 and v8, at same scales, however when switch to the DOTAv1.5 dataset, we see a different story, for small scale images v8 outperforms the next best model (v11) by nearly 4 points and the gap only widens when considering the rest. For medium-scale objects, we see that v9 leads by a large margin of 9 points compared to the next best v8 model. And for large objects we see the widest margin in our testing, where v8 outperforms all the other models. From our testing, we clearly see that while the overall performance of the models is similar (especially when comparing the overall performance of v8 and v9) , when it comes to dealing with objects at different scales, these models display wildly different behaviors. 

\begin{figure}
    \centering
    \includegraphics[width=1\linewidth]{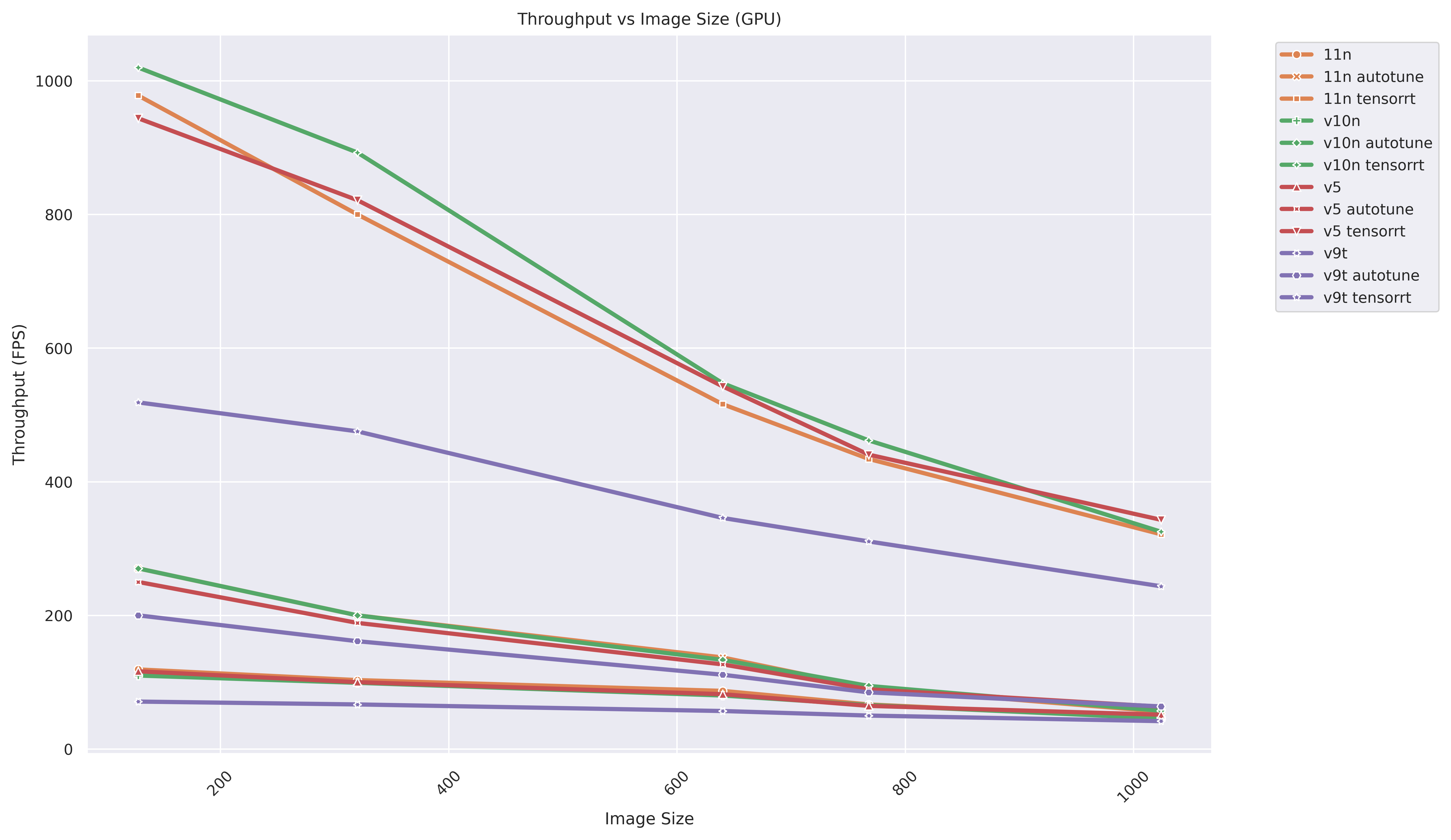}
    \caption{Effect of image size on throughput of GPU}
    \label{fig:imagesize gpu}
\end{figure}
\subsection{Small Object Performance Analysis}
The 4-point lead of YOLOv8 over YOLOv11 for small objects in DOTAv1.5 suggests that YOLOv8's design maintains finer spatial resolution in its feature maps, critical for aerial imagery where objects like vehicles and small buildings appear at diminished scales. This performance difference can be attributed to YOLOv8's C2f module, which effectively preserves spatial information while reducing computational complexity \cite{lightyolov8}.
Furthermore, YOLOv9's substantial 9-point advantage in medium-sized object detection can be linked to its Programmable Gradient Information (PGI) approach, which reduces information bottlenecks that typically occur in deep neural networks \cite{yolov9}. This improvement contributes to the model performing second best on the DOTAv1.5 dataset. The comparatively lower mAP scores of v10 and v11 on the DOTAv1.5 dataset despite their great performance on the COCO dataset might suggest that the attention mechanism for these models performs better on low instances of small objects, as the DOTAv1.5 dataset has hundreds of small objects per image, which may overwhelm the attention mechanism.
\begin{figure}
    \centering
    \includegraphics[width=1\linewidth]{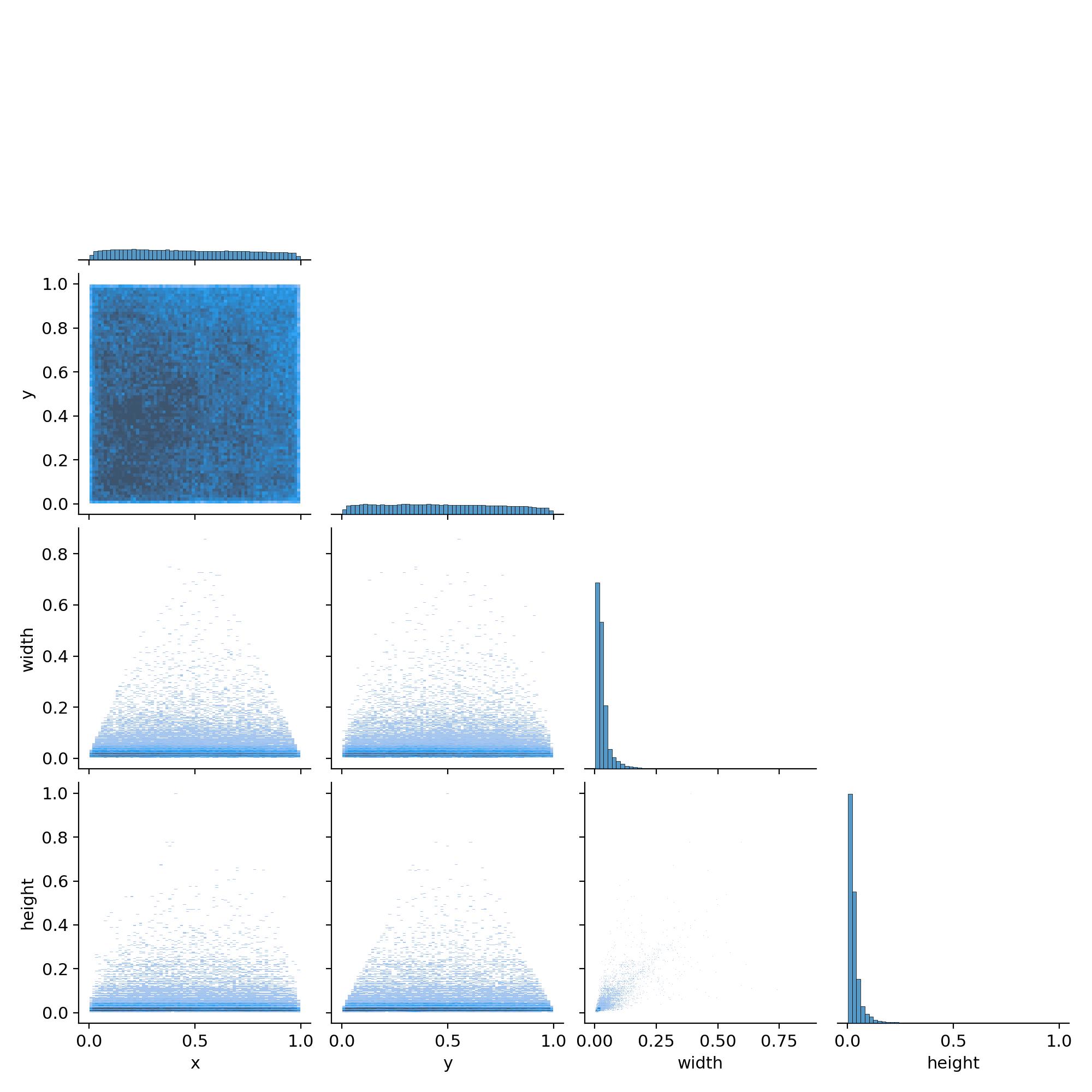}
    \caption{DOTAv1.5 dataset label parallelogram}
    \label{fig:labels_correlogram}
\end{figure}

\section{Conclusion}
In this paper, we have performed an extensive evaluation of multiple YOLO models. Their accuracies, latency, and throughput at various image sizes and their performance using the wide variety of available software backends. Here we found that openvino showed the best performance on both the Intel and AMD platforms. While on the GPU side of things, Tensorrt outperformed all other backends by an extremely wide margin. We also did an evaluation of their accuracy in the detection of small objects at 1\%, 2.5\% and 5\% of the overall image for the COCO and DOTAv1.5 datasets, where we find the v8 and v9 models showing great performance.

\bibliographystyle{ieeetr}
\bibliography{refs.bib}

\end{document}